\documentclass[prl,aps,floatfix,superscriptaddress,noshowpacs,twocolumn]{revtex4}
\usepackage[mathlines]{lineno}

\usepackage{dsfont}
\usepackage{amsmath,amssymb,graphicx,bm,color,mathrsfs,verbatim,epstopdf,dcolumn,cancel}
\usepackage{booktabs}
\usepackage{siunitx}

\usepackage{hyperref}
\hypersetup{ 
	colorlinks   =  true
}

\graphicspath{
	{paper_figs/linearseperable/},
	{paper_figs/nonlinearseperable/}
	{paper_figs/mnist/}
	{paper_figs/graphic/}
	{paper_figs/SVDD/}
}

\begin{document}

\title{Machine Learning as Ecology}

\author{Owen Howell}
\email{olh20@bu.edu}
\affiliation{Department of Physics, Boston University, 590 Commonwealth Ave., Boston, MA 02215, USA}

\author{Cui Wenping}
\affiliation{Department of Physics, Boston University, 590 Commonwealth Ave., Boston, MA 02215, USA}
\affiliation{Department of Physics, Boston College, 140 Commonwealth Avenue, Chestnut Hill, MA 02467}

\author{Robert Marsland III}
\affiliation{Department of Physics, Boston University, 590 Commonwealth Ave., Boston, MA 02215, USA}

\author{Pankaj Mehta}
\email{pankajm@bu.edu}
\affiliation{Department of Physics, Boston University, 590 Commonwealth Ave., Boston, MA 02215, USA}

\date{\today}


\begin{abstract}
Machine learning methods have had spectacular success on numerous problems. Here we show that a prominent class of learning algorithms - including Support Vector Machines (SVMs) -- have a natural interpretation in terms of ecological dynamics. We use these ideas to design new online SVM  algorithms that exploit ecological invasions, and benchmark performance using the MNIST dataset.  Our work provides a new ecological lens through which we can view statistical learning and opens the possibility of designing ecosystems for machine learning. 
\end{abstract}

\maketitle

\section{Introduction}

Machine learning (ML) is one of the most exciting and useful areas of modern computer science \cite{bishop2006pattern, mehta2019high}. One common machine learning task is classification: given labeled data from one or more categories, predict the category of a new, unlabeled data point. Another common task is to perform outlier detection (i.e. find data points that appear to be irregular). Both of these difficult problems can be solved efficiently using kernel-based methods such as Support Vector Machines (SVMs) \cite{cortes1995support,  scholkopf2002learning,bishop2006pattern}. 

\begin{figure}[t]
	
	\includegraphics[width=1.0\columnwidth]{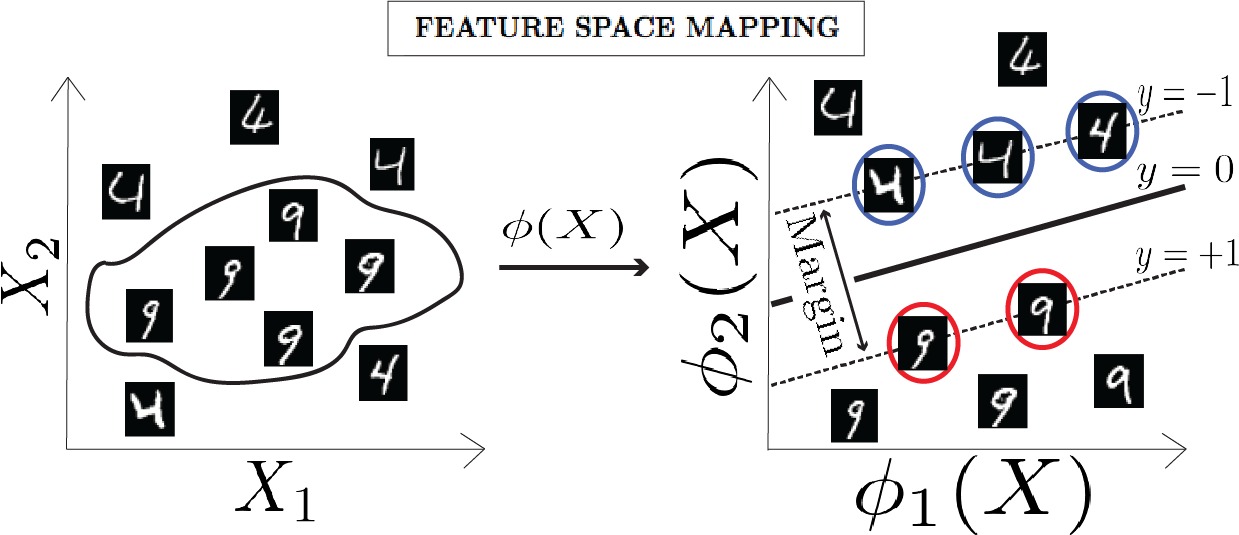}
	\caption{Overview of Support Vector Machines (SVMs). Data points are mapped into a high-dimensional feature space via $\phi(X)$ where they can be separated using a linear decision
	surface. The SVM tries to maximize the distance (margin) from the decision boundary to the nearest data point. Points that lie on the maximum-margin planes  (circled) are called support vectors and used
	to classify new, unlabeled data. } 
	\label{Feature_space_transform}
	
\end{figure}

The basic idea behind SVMs is to use a non-linear map to embed the input data in a high-dimensional feature space where it can be classified using a simple linear classifier (see Figure \ref{Feature_space_transform}). To ensure good generalization and avoid overfitting, SVMs focus on the ``hardest to classify'' points that lie closest to the linear decision surface in the high-dimensional feature space. These points are called ``support vectors'' and play a prominent role in SVM algorithms.

The real power and utility of SVMs comes from the fact that these ideas can be implemented quickly and efficiently using kernel methods and quadratic optimization \cite{scholkopf2002learning,bishop2006pattern}. The idea of a kernel function is to replace the \emph{explicit} mapping to a high-dimensional feature space with an \emph{implicit} kernel function that specifies the similarity (dot product) between data points in the high-dimensional feature space. Once the kernel function is specified, the support vectors and decision surface can be easily computed as an instance of a Quadratic Programming (QP) problem. There exist efficient exact and approximate optimization algorithms for QP that scale weakly polynomially in input size.

The original motivation for SVMs and other kernel methods were deep results in statistical learning theory concerning generalization errors \cite{vapnik2013nature, scholkopf2002learning, cortes1995support}. Here, we show that these statistical problems can also be understood using ideas from niche theory in community ecology (see Table \ref{ecotable}) \cite{Mac_1,Chesson_1}. Our construction exploits the recently discovered duality between ecological dynamics and constrained optimization problems, specifically quadratic programming \cite{Mehta_18,Levin_10,Mehta_17}. In particular, we show that data points can be viewed as ``species'' that compete for resources, with each feature identified with a distinct resource, and the kernel function specifying the niche overlap between species/datapoints \cite{macarthur1967limiting, colwell1971measurement}. 

This mapping allows us to reinterpret SVMs as complex ecosystems that self-organize into ecologically stable steady states defined by their support vectors.  This new ecological perspective naturally leads to a new \emph{online} algorithms based on ecological invasion for SVMs as well as for outlier detection kernel methods such as Support Vector Data Description (SVDD)\cite{Scholkopf_1999,Tax_2004}. We also show that  our ecological SVDD method is equivalent to the online algorithm derived in \cite{SAS_18}.

\section{ SVMs as QP}
Consider a classification problem where each $p$-dimensional data point $x_i$ ($i = 1,2,3\dots N)$ comes with a binary label $t_i=\pm 1$.
A SVM fits a linear classifier to the data of the form 
\begin{equation}
y(x)  = w^{T}\phi(x) + b
\end{equation}
where $\phi : \mathbb{R}^{p} \to \mathbb{R}^{q}  $, $q \gg p$ denotes a mapping to a high-dimensional feature space. The scalar offset $b$ and the $q$-dimensional weight vector $w$ are tunable model parameters. 

 A new data point $x_k$ is assigned to class $t_k=+1$ if $y(x_k)>0$ and to class $t_k=-1$ if $y(x_k)< 0$. In the main text, we restrict our discussion to linearly separable datasets, i.e., datasets for which exists a hyperplane in the feature space  $\phi(x)$ that partitions the dataset into two regions with every point in class $+1$ in one region and every point in class $-1$ in the other (see Fig \ref{Feature_space_transform}). However, our construction can be easily generalized to non-separable datasets (see Supporting Information). 

SVMs are trained by maximizing the margin, defined as the Euclidean distance from the line $y(x) = 0 $ (the decision boundary) to the nearest data point. It is easy to show that distance from the point $x_i$ to the line $y(x)=0$ is given by the expression $t_{i} \frac{y(x_i)}{|w|} $. Maximizing the margin corresponds to choosing the parameters  $w$ and $b$ so that
\begin{align}\label{SVMmodelmain}
\begin{split}
w,b =&\text{ arg max}_{w,b}\left\{ \frac{1}{|w|} \text{min}_{i} [t_{i} (w^{T}\phi(x_{i}) + b)  ] \right\}
\end{split}
\end{align}
The above maximization problem can be recast by noting that Equation \eqref{SVMmodelmain} has a gauge degree of freedom: the decision surface is invariant under the scaling transformation $w \rightarrow D w$ and $ b \rightarrow D b$  \cite{ cortes1995support, scholkopf2002learning,bishop2006pattern}. We can fix this gauge by choosing the margin to be exactly 1. In this gauge, Equation \eqref{SVMmodelmain}  is equivalent to the following convex quadratic programming problem 
\begin{align}\label{primalmain} 
\begin{split}
&\text{arg min}_{w,b} \frac{1}{2} |w|^{2} \\
& \text{subject to } t_{i} ( w^{T}\phi(x_{i}) +b ) \geq 1 \text{ for all }i,
\end{split}
\end{align}
where $i$ labels the $N$ data points in the training dataset.

As with all constrained optimization problems, we can also solve the equivalent  dual optimization problem by introducing  generalized Lagrange multipliers $a_i$ (often called KKT multipliers in the optimization literature) corresponding to each of the inequality constraints in \eqref{primalmain} \cite{Boyd}. Since there is one constraint per data point $i$, we can uniquely associate each $a_i$ with a  data point in the training set. For data points that saturate the inequality in (\ref{primalmain}), $a_i$ is positive, and acts as an ordinary Lagrange multiplier to enforce the constraint. For the rest of the data points, no Lagrange multiplier is required, and $a_i=0$. These observations give rise to the Karush-Kuhn-Tucker conditions, which are necessary and sufficient to determine the optimum \cite{ cortes1995support, scholkopf2002learning,bishop2006pattern}:
\begin{align}
\begin{split}
0&=\nabla_{w,b} L(w,b,a_i)\\
1 &\leq t_{i} ( w^{T}\phi(x_{i}) +b )\\
0 &\leq a_i\\
0 &= a_i[t_{i}(w^{T}\phi(x_{i})+ b) -1 ]
\end{split}
\end{align}
where the last three expressions hold for all $i$, with the SVM Lagrangian
\begin{align}
\label{SVMlag1main}
L(w,b,a_{i}) = \frac{1}{2} |w|^{2} - \sum_{i=1}^{N} a_{i} [t_{i}(w^{T}\phi(x_{i})+ b) -1 ].
\end{align}

Solving the first condition for $w$ and $b$ yields the equations $w = \sum_{i=1}^{N} a_{i}t_{i} \phi(x_{i})$ and $\sum_{i=1}^{N} a_{i}t_{i} = 0$.
Inserting these results into Equation \eqref{SVMlag1main} gives equations for optimal $a_{i}$: 
\begin{align}\label{lagmain}
\begin{split}
&\text{argmax}_{a_{i}} \text{ } L(a_{i}) = \sum_{i=1}^{N} a_{i} - \frac{1}{2} \sum_{i,j=1}^{N} a_{i}a_{j}t_{i}t_{j} K(x_{i},x_{j}) \\ 
& \text{subject to }0 \leq a_{i} \text{ for all }i  \\
&\text{and } \sum_{i=1}^{N}  a_{i}t_{i}  = 1
\end{split}
\end{align}
$L(a_{i})$ is called the dual SVM Lagrangian. In writing this equation, we have introduced the kernel function $K(x_{i},x_{j}) \equiv \phi^{T}(x_{i}) \phi(x_{j})$ which is just the dot product of the data points in the high-dimensional feature space $\phi$.

In this dual formulation, the support vectors correspond precisely to those data points $x_k$ for which the corresponding KKT multiplier is greater than zero $a_k> 0$. The SVM can be used to classify a new point $x$ using $t=\text{sign}(y(x))$ with
\begin{align*}
y(x) &= \sum_{i\in S} t_{i} a_{i} K(x,x_{i}) + b\\
b &= \frac{1}{|S|} \sum_{i \in S} \left[ t_{i} - \sum_{j \in S} a_{j}t_{j}K(x_{i},x_{j}) \right]
\end{align*}
and $S$ the set of support vectors.

\section{The Ecology of SVMs}

 Consider the maximization of the dual Lagrangian $L(a_i)$ given in Equation \eqref{lagmain}, subject to the constraints $\sum_{i=1}^{N} a_{i}t_{i} = 0$ and $a_i \geq 0$.  Recently, it was shown there exists a duality between constrained optimization and ecological dynamics \cite{Mehta_18}. Using this duality, it is straightforward to show that the solution to this problem is encoded in the steady state of a generalized Lotka-Volterra equation of the form 
\begin{align}\label{EcologicalODEmain}
\begin{split}
&\frac{d a_{i}}{dt} = a_{i} \left[1 + \lambda t_{i} - \sum_{j=1}^{N} t_{i} t_{j} K(x_{i},x_{j})  a_{j} \right] \\
&\frac{d \lambda}{dt} = -\sum_{i=1}^{N} a_{i} t_{i} ,
\end{split}
\end{align}
This system of differential equations has a natural ecological interpretation as the dynamics of $N$ species with abundances $a_{i}$ ($i=1\ldots N$) whose interactions are represented by the matrix $\alpha_{ij}$ with elements
\begin{equation}
\alpha_{ij}=t_{i}t_{j} K(x_{i},x_{j}).
\end{equation}
Since each $a_i$ corresponds to a data point, we can think of this as an ecological network where data points $i$ and $j$ from the same class ( $t_i=t_j$) compete with each other (i.e. $\alpha_{ij}>0$) whereas species
of from different classes ($t_i = -t_j$) are mutualistic (i.e. $\alpha_{ij}<0$). The level of competition or mutualism depends on the overlap kernel $K(x_i, x_j)$ with similar data points having stronger interactions.

Ecologically, a combination of competitive and mutualistic interactions such as these naturally occur in plant-pollinator networks \cite{Valdovinos_Adaptive_Foraging}. In such networks, different species of plants compete with each other for pollinators, pollinators compete with each other for plants, and  plant-pollinators
interactions are beneficial for both kinds of species. The $\lambda$ term corresponds to an abiotic environmental factor that is produced or consumed by different species. In this plant-pollinator analogy, $\lambda$ could represent an environmental $\text{CO}_{2}$ concentration. Specifically, plants consume $\text{CO}_{2}$ and benefit from high $\text{CO}_{2}$ concentration while pollinators produce  $\text{CO}_{2}$ and are harmed by high $\text{CO}_{2}$ concentration. 

Note that this interpretation differs from the consumer-resource interpretation given to a generic constrained optimization problem in \cite{Mehta_18}. The Lagrange multiplier $\lambda$ plays the role of a ``resource,'' but is not required to be positive, since it is enforcing an equality constraint rather than an inequality. The variables $a_i$ of the optimization are now treated as the species rather than as resources.

\begin{table}

	\centering
	\begin{tabular}{| l || r | }
		\hline
		{\bf SVM}&{\bf Ecology}\\
		\hline 
		Data point &Species\\
		KKT Multiplier & Species Abundance\\
		Feature Space & Trait Space\\
		Kernel & Niche Overlap\\
		Support Vectors & Species that survive in ecosystem\\
		\hline

	\end{tabular}
	\caption{Conceptual mapping between SVMs and ecology 				\label{ecotable}	 }

\end{table}

These observations suggest a new ecological interpretation of SVMs (see Table \ref{ecotable}). Data points act like species that either compete or promote each others' survival. The abundance of each species is the value of KKT multiplier that enforces the corresponding constraint  in \eqref{primalmain}. Since only the support vectors have non-zero KKT multipliers, the only data points that survive in the ecosystem are support vectors. As noted above, data points from  the same category compete with each whereas data points from different categories are mutualistic. As is widely appreciated in the ecological literature, the ecological dynamics depends only on the overlap of resource utilization function encoded in the similarity kernel $K(x_i, x_j)$ between points \cite{macarthur1967limiting}. The data points most likely to survive in the ecosystem are data points from one category  that are similar to data points from the opposite category since they have large mutualistic interactions.  For this reason, the data points that survive in the ecosystem are precisely those lie near the boundary between the two categories, that is, the support vectors. 

This mapping can also be easily generalized to the case where the data is not linearly separable, and to Support Vector Data Description (SVDD) algorithms \cite{Scholkopf_1999,Tax_2004,SAS_18} for outlier detection (see SI).

 \section{EcoSVM: An Online Algorithm} 
 
 One interesting class of processes that has been extensively studied in the ecological literature is ecological invasion \cite{Levin_10,Tilman_10,Chesson_2,Case_1990}. In the context of SVMs, invasion by new species corresponds to addition of a new data point $(x_{0},t_{0})$ to our existing dataset.  If we denote the existing support vectors by the set $S$, the condition for a successful invasion is the intuitive statement that the initial growth rate must be positive when the new
data point is introduced into the ecosystem:
\begin{equation}\label{growthconditionmain}
0 < \frac{1}{a_0} \frac{da_{0}}{dt} = 1 + \lambda t_{0} - \sum_{j \in S}  t_{0}t_{j} K(x_{0},x_{j}) a_{j} .
\end{equation}
When this equation is satisfied the new data point can successfully invade the ecosystem and fixate (i.e. become a support vector).  If the condition is not satisfied, the point goes ``extinct'' and the set of support vectors does not change. If a data point can invade successfully, the species abundances ``$a_i$''  are modified and can be found by solving for the steady state of \eqref{EcologicalODEmain} using either forward integration or quadratic programming \cite{Mehta_18}. 

This suggests a simple new approximate algorithm for online SVM learning we term the EcoSVM. In online learning, rather than seeing all the data at once, training data is presented in a sequential pattern.  In the EcoSVM algorithm when a new training data point is presented,  the invasion condition \eqref{growthconditionmain} is used to determine whether it can successfully invade the ecosystem. If it cannot, the training data point is discarded. If it can, we recompute the steady-states using  Equation \eqref{EcologicalODEmain}.  This algorithm can be easily generalized to the case of non-separable data (see Supporting Information and attached code implementing the algorithm). 

Because we use the ecologically inspired invasion condition, there is no need to recompute the support vectors at each learning step, resulting in a faster and more memory-efficient online algorithm than those that were previously suggested \cite{Poggio_10,Langfod_2011,Laskov_2006,Tax+Laskov_2003,Ruff_18,Sohrab_18}. The EcoSVM algorithm also reduces the amount of training data that needs to be stored in memory. Specifically, instead of needing to store all data points, we keep only the support vectors. Since the number of support vectors is in general a small subset of all the training data, this greatly reduces the memory requirements. This increased efficiency comes at the expense of introducing small errors that come from the contingent nature of ecological invasions. Occasionally, a successful invasion by a new species (new data point) will allow a species that could not previously invade the ecosystem (designated not a support vector) to become viable (a support vector). This kind of historical contingency introduces errors in our online algorithm since we discard all data points that do not fix in the ecosystem. In practice, we find that these errors are generally quite small for real-world datasets.

\section{Numerical Tests}
 
\begin{figure}[t!]
	\includegraphics[width=0.49\columnwidth]{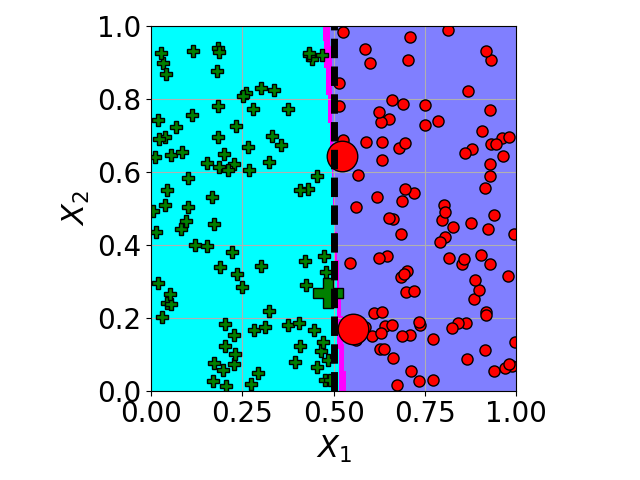}
	\includegraphics[width=0.49\columnwidth]{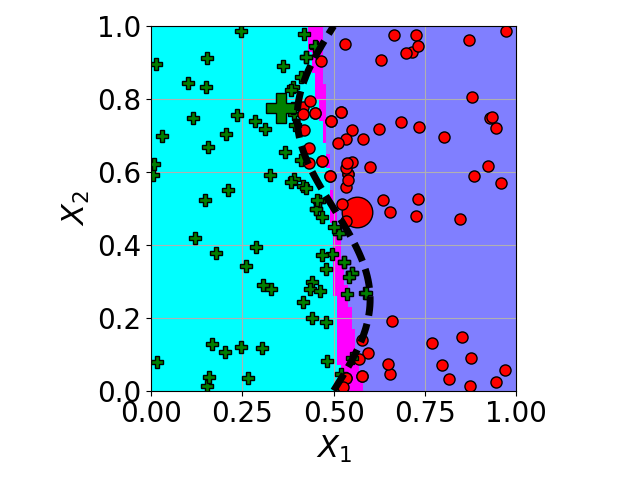}
	\caption{Comparison of classical SVM and EcoSVM algorithms for a data set with $N=200$ total training points points for  (Left) a linearly separable dataset and  (Right) a dataset where 
	the data is not linearly separable (see SI for more detail on algorithm in non-linear case).  The true decision boundary is given by dashed black line. Cyan regions show data that the full SVM and online SVM both identify as $t_{i}=1$. Blue regions show data that full SVM and online SVM both identify as $t_{i}=-1$. Purple regions show area in which SVM and online SVM disagree. $t_{i}=1$ data points are shown as green plus symbols, $t_{i}=-1$ data points are shown as red circles. Active support vectors are shown with larger symbols.} 
\label{linsepfig}
\end{figure}

 We tested EcoSVM using several numerical examples. We started with two toy datasets with $N=200$ data points $x=(c_1,c_2)$ drawn uniformly from a two-dimensional hypercube $ x \in [0,1]^{2}$. We enforced linear separability using a decision boundary given by $c_1=1/2$ so that data points with $c_1 <1/2$ were assigned to one class $t=-1$ and those with $c_1>1/2$ were assigned to a different class $t=1$ (see Figure \ref{linsepfig}). We also created a dataset where the decision boundary was given by $c_{1} = \frac{1}{2} + \frac{1}{10} \sin(2\pi c_{2})$, which is not linearly separable in the given feature space. After initialization using the first 10 data points, we trained an EcoSVM using the online scheme described above. As can be seen visually in Figure \ref{linsepfig}, the decision boundaries found by the EcoSVM were very similar to those found using an ordinary batch SVM algorithm. Furthermore, the final accuracy and number of support vectors of the EcoSVM algorithms and ordinary batch SVM algorithm were almost identical (See SI for more detail).

\begin{figure}[t!]
	\begin{tabular}{ll}
		\includegraphics[width=0.5\columnwidth]{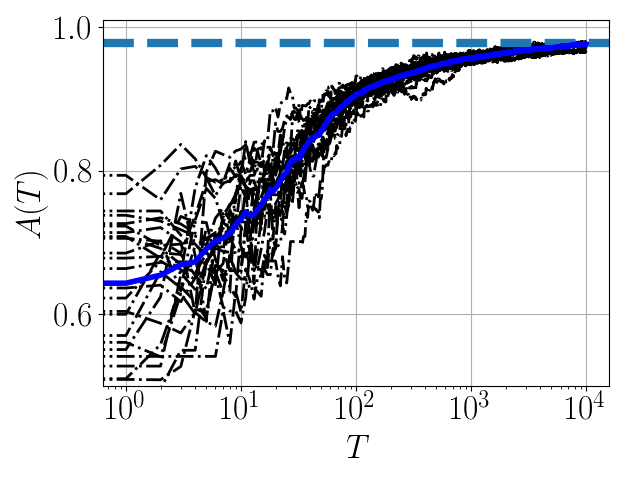}
		\includegraphics[width=0.5\columnwidth]{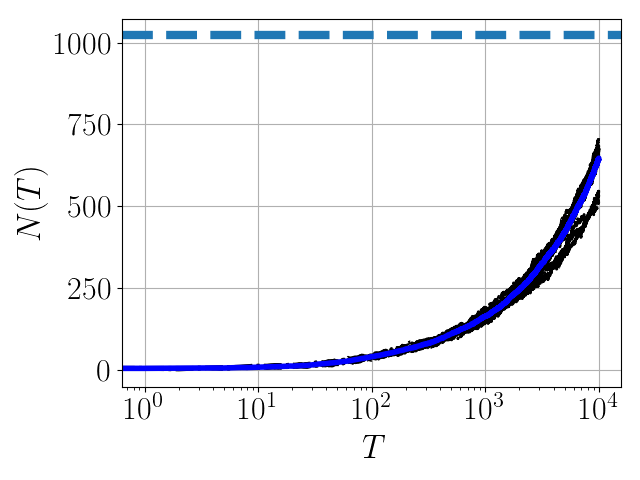}
	\end{tabular}	
	\caption{An ecologically inspired online SVM algorithm EcoSVM applied to digit classification of ones and fours from the MNIST dataset \cite{MNISTDATA,Lecun98gradient-basedlearning}. Accuracy of EcoSVM  for 25 different realizations. The average accuracy over all realizations is shown using the solid blue line and the accuracy of an SVM trained on the entire dataset is shown using the blue dotted line. The inset panel shows the number of active support vectors in each realization (black lines), the mean number of support vectors across all realizations (blue solid line) and the number of support vectors for SVM trained on full dataset (dotted blue line). }   
	\label{mnistfigacc}
\end{figure}

Next, we tested the performance of EcoSVM algorithm on MNIST \cite{MNISTDATA,Lecun98gradient-basedlearning}, a standard benchmark dataset in machine learning. The MNIST dataset consists of $6,000$ training images and $1,000$ test images of each of the handwritten digits `0'-`9'. To test the EcoSVM, we considered the binary classification task of distinguishing fours and nines. For this classification problem, we used a standard Gaussian (RBF) kernel given by $K_{\sigma}(x,y) = \exp\left[-\frac{1}{2 \sigma^{2}} (x - y)^{2}\right]$, where the kernel width $\sigma$ was determined via cross validation on the batch SVM. The performance of our EcoSVM algorithm was comparable to a traditional SVM trained on the full dataset ( $98.1 \%$ accuracy compared to  $98.5\%$ accuracy for traditional SVMs, as shown in Figure \ref{mnistfigacc}). The EcoSVM algorithm also ends up finding a similar number of support vectors as a traditional SVM: $\sim 750$. We note that since the MNIST dataset is not completely linearly separable in the RBF feature space, in these numerical simulations we used the generalization of the EcoSVM algorithm for non-linearly separable datasets discussed in the SI.

 \section{Conclusion}
 
 In this work, we have shown how we can think about kernel methods using ideas from ecology. This ecological mapping allowed us to formulate a new ecologically inspired online SVM algorithm, the EcoSVM. We have shown that the performance of EcoSVM is comparable to traditional SVMs that work with all data simultaneously. Our algorithm differs from previous online SVMs which must recompute the support vectors at each learning step \cite{Poggio_10,Langfod_2011,Laskov_2006,Tax+Laskov_2003,Ruff_18,Sohrab_18} allowing for faster implementation and smaller memory requirements. While in the main text we focus on  linearly separable data, as shown in the SI these same ideas can be generalized to non-separable data and for outlier detection (SVDD) using unlabeled data. Our results suggest that ecological dynamics may provide a rich new setting for thinking about biologically-inspired machine learning. They also suggest the tantalizing possibility that it maybe possible to engineer synthetic ecosystems that implement sophisticated statistical learning algorithms \cite{zomorrodi2016synthetic}.

\section{Acknowledgments}

OH acknowledges support from BU UROP student funding. The work was supported by NIH NIGMS grant 1R35GM119461, Simons Investigator in the Mathematical Modeling of Living Systems (MMLS) to PM.

\newpage

\section*{\large Supporting Information}

Code implementing the EcoSVM algorithm is found at https://github.com/owenhowell20/EcoSVM.

\section{Adding The Slack}

In the main text we have focused on datasets that are linearly separable. For the majority of practical applications this is not the case. For overlapping class distributions, the primal SVM problem is modified so that points are allowed to be on the wrong side of the margin. Specifically, slack variables $\zeta_{i} \geq 0 $ are introduced with $t_{i} y(x_{i}) \geq 1- \zeta_{i}$. This should be compared with the linearly separable case where the constraint is instead $ t_{i} y(x_{i}) \geq 1$. The new minimization is weighted to penalize points that lie on the wrong side of the margin
\begin{align}\label{primalslack} 
\begin{split}
&\text{arg min}_{w,b,\zeta_i}\,\,\, \frac{1}{2} |w|^{2} + C\sum_{i=1}^N \zeta_{i}\\
& \text{subject to } t_{i} ( w^{T}\phi(x_{i}) +b ) \geq 1 \text{ for all }i,
\end{split}
\end{align}
where the slack parameter $C$ determines the extent to which points on the wrong side of the margin are tolerated. In practice, $C$ is a hyper-parameter that is tuned to minimize generalization error. The KKT conditions for this new minimization problem are:
\begin{align}
\begin{split}
0&=\nabla_{w,b,\zeta_i} L(w,b,\zeta_i, a_i, \mu_i)\\
1  - \zeta_{i} &\leq t_{i} y(x_{i})\\
0 &\leq a_i\\
0 &= a_i[t_{i}y(x_{i}) -1 + \zeta_{i} ]\\
0 &\leq \zeta_i\\
0 &\leq \mu_i\\
0 &= \mu_i \zeta_i
\end{split}
\end{align}
where the $\mu_i$ are additional KKT multipliers enforcing the constraints $\zeta_i \geq 0$, and the primal Lagrangian is

\begin{align*}
\begin{split}
\label{SlackLag}
& L(w,b,\zeta_{i},a_i,\mu_i)  = \\ & \frac{1}{2} |w|^{2} +  \sum_{i=1}^{N}  \left[C \zeta_{i} - a_{i}( t_{i} y(x_{i}) - 1 + \zeta_{i} )-\mu_i\zeta_i\right]
\end{split}
\end{align*}

Minimizing the Lagrangian $ \frac{\partial L }{\partial w_{i}} = 0$ ,$\frac{\partial L }{\partial b} = 0$ and $\frac{\partial L }{\partial \zeta_{i}} = 0$ gives equations 
\begin{equation}
w = \sum_{i=1}^{N} t_{i} a_{i} \phi(x_{i})  \text{ , } 	\sum_{i=1}^{N} a_{i} t_{i} = 0  \text{ , }	a_{i} = C - \mu_{i}
\end{equation}
Each $\mu_{i} \geq 0$ so the last equation is equivalent to $ a_{i} \leq C$. Inserting these results into the primal Lagrangian transforms the problem into maximization of the dual SVM Lagrangian 

\begin{align}\label{Slacklag}
\begin{split}
&\text{argmax}_{a_{i}} \text{ } L(a_{i}) = \sum_{i=1}^{N} a_{i} - \frac{1}{2} \sum_{i,j=1}^{N} a_{i}a_{j}t_{i}t_{j} K(x_{i},x_{j}) \\ 
& \text{subject to }0 \leq a_{i} \leq C \text{ for all }i  \\
&\text{and } \sum_{i=1}^{N}  a_{i}t_{i}  = 1
\end{split}
\end{align}

We can enforce the second constraint by introducing a Lagrange multiplier $\lambda$, resulting in the following set of equations for the optimal $a_i$:

\begin{align}\label{lag2}
\begin{split}
&\text{argmax}_{a_{i},\lambda} \text{ } L(a_{i},\lambda) \\ 
& \text{subject to }0 \leq a_{i} \leq C \text{ for all }i  \\
\end{split}
\end{align}
with Lagrangian
\begin{equation}
L(a_{i},\lambda) = \sum_{i=1}^{N} a_{i} - \frac{1}{2} \sum_{i,j=1}^{N} a_{i}a_{j}t_{i}t_{j}K(x_{i},x_{j}) +  \lambda \sum_{i=1}^{N}  t_{i}a_{i}
\end{equation}

Using the duality described in \cite{Mehta_18}, we can map the quadratic programming problem \eqref{lag2} to ecological dynamics
\begin{align}\label{EcologicalODEslack}
\begin{split}
&\frac{d a_{i}}{dt} = a_{i} (C - a_{i}) (  1 + \lambda t_{i} - \sum_{j=1}^{N} t_{i} t_{j} K(x_{i},x_{j})  a_{j} ) \\
&\frac{d \lambda}{dt} = -\sum_{i=1}^{N} a_{i} t_{i}.
\end{split}
\end{align}
where the prefactor $a_i(C-a_i)$ enforces the constraints on $a_i$. Equation \eqref{EcologicalODEslack} has a similar interpretation to the Lotka-Volterra equations for the linearly separable case, with the additional $(C-a_{i})$ factor can be interpreted as each species having a maximum carrying capacity $C$ \cite{Valdovinos}.

Now consider the addition of new point $P_{0}=(x_{0},t_{0})$. This point changes the set of support vectors if the initial growth rate is positive.
\begin{equation}\label{growthconditionslack}
0 < \frac{1}{a_{0}} \frac{da_{0}}{dt} = 1 + \lambda t_{0} - \sum_{j =1}^{N}  t_{0}t_{j} K(x_{0},x_{j}) a_{j}
\end{equation}
Let $x_{k}$ be any ``active'' support vector, that is, a point whose KKT multiplier $a_k$ satisfies $C>a_{k}>0$. Then, solving the steady state equation \eqref{EcologicalODEslack} for auxiliary variable $\lambda$ gives
\begin{equation}\label{SVMlambda}
\lambda = -t_{k} + \sum_{i=1}^{N} t_{i} K(x_{i},x_{k})a_{i} 
\end{equation}
Inserting this into Equation \eqref{growthconditionslack} gives the invasion condition
\begin{equation}\label{growthconditionslack2}
0 < \frac{1}{a_{0}} \frac{da_{0}}{dt} = 1 - t_{k}t_{0} + \sum_{i=1}^{N} t_{i}t_{0} ( K(x_{i},x_{k}) - K( x_{i} , x_{0}  )  ) a_{i} 
\end{equation}
 This invasion condition can be used to construct an online learning algorithm. Specifically, when a new data point is presented, the condition \eqref{growthconditionslack2} can be used to determine whether the new point changes the set of support vectors without having to recompute the minimum of \eqref{Slacklag}. The nonzero KKT multipliers $a_{i}>0$ and corresponding support vectors $x_{i}$ are kept in memory.

A new point $x$ is classified using $t=\text{sign}(y(x))$ with
\begin{align*}
y(x) &= \sum_i t_{i} a_{i} K(x,x_{i}) + b\\
b = &\frac{1}{|M|} \sum_{i \in M} \left[ t_{i}   - \sum_{j \in S} a_{j}t_{j}K(x_{i},x_{j})  \right]
\end{align*}
where $S$ is the full set of support vectors and $M$ is the subset of active support vectors. Note that this formula requires at least one active support vector.

\section{Performance on Toy Models}

 We test our proposed online learning algorithms on two toy datasets. We consider one dataset that is linearly separable in the feature space $\phi(x) = x $. Specifically, we choose all data points to be drawn from the $[0,1]^{p}$ $p$-dimensional hypercube. We then define the decision surface:
\begin{equation*}
B_{1} : (x_{1}=\frac{1}{2},x_{2},...x_{p}) 
\end{equation*}

We consider a second dataset that is not linearly separable. Specifically, we define the second dataset to have decision boundary given by: 
\begin{equation*}
B_{2} : (x_{1} = \frac{1}{2} + \frac{1}{10} \sin(2\pi x_{2})\sin(2\pi x_{3})...\sin(2\pi x_{p}), x_{2},...x_{p}) 
\end{equation*}

To test our proposed algorithm, we draw $N$ points from the $p$ dimensional hypercube. The minimum of the SVM Lagrangian is found for $N_{s}$ points with $N_{s} \ll N$. We require that $N_{s}$ is greater then $p$ or else there are flat directions and our algorithm can become unstable. At each step, a new point is presented and the invasion condition \eqref{growthconditionslack} is used to determine whether the set of support vectors is changed. If \eqref{growthconditionslack} is satisfied, the steady state is recomputed using quadratic programming. This is continued for all $N$ points. We find an excellent agreement between the predictions of our online algorithm and an batch SVM trained using all $N$ points for both the linearly separable and non-linearly separable datasets.

We study how the test accuracy and number of support vectors depend on the training epoch $T$. For this purpose, let us define the accuracy
\begin{equation*}
A(T) =  1 - \frac{1}{|N_{test}|} \sum_{x\in N_{test} } \frac{1}{2} | t_{T}(x) - t_{Exact}(x)|
\end{equation*}
where $N_{test}$ is the set of testing data and $t_{Exact}(x)$ is the true label corresponding to point $x$. $t_{T}(x)$ denotes the prediction of the online SVM trained with $T$ data points. In addition, let us define 
\begin{equation*}
N(T) = \begin{cases}
\text{Number of } a_{i}(T) > 0 \\ \text{ for linearly seperable case} \\
\text{Number of } C > a_{i}(T) > 0\\ \text{ for non-linearly seperable case} 
\end{cases}
\end{equation*}
where $a_{i}(T)$ are the support vector coefficients of the online SVM trained with $T$ data points. In both the linear and non-linear case $N(T)$ counts the number of active support vectors.

\begin{figure}[ht]

	\begin{tabular}{ll}
		\includegraphics[width=0.5\linewidth]{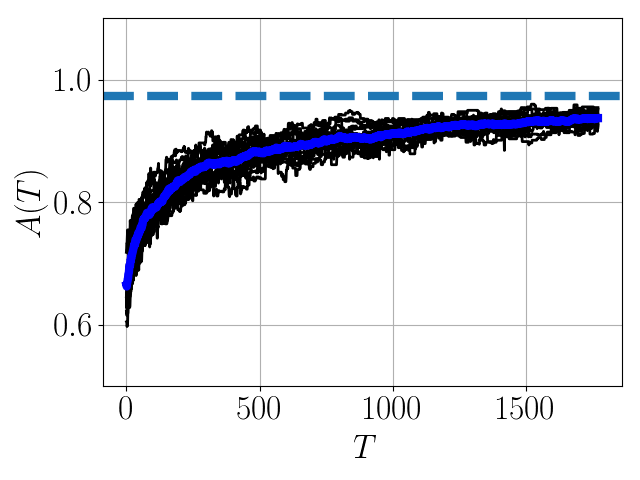}
		\includegraphics[width=0.5\linewidth]{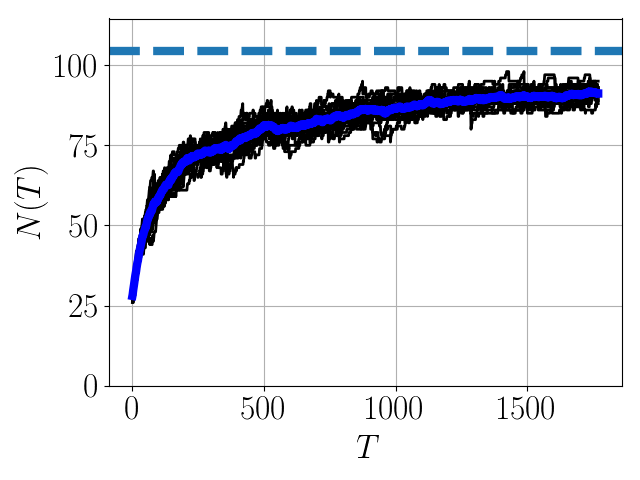}
	\end{tabular}

	\caption{Test accuracy and number of support vectors as a function of time for the linearly separable toy model, with true decision boundary $B_1$ defined in the SI text. Left panel shows accuracy of online SVM algorithm $ A(T) $ as a function of the number of points $T$ that the online SVM has seen. Black lines show individual realization, blue line shows mean accuracy. Dotted blue line shows full SVM accuracy.   Right panel shows the number of support vectors $N(T)$ as a function of the number of points $T$. Black lines show individual realization, blue line shows mean number of support vectors. Dotted blue line shows number of support vectors in SVM trained on entire dataset at once. The dimension of the data space is $p=100$ and the online training is initialized with $N_{s}=30$ data points.}
	\label{linearFIG}
\end{figure}

\begin{figure}[h!]

	\begin{tabular}{ll}
		\includegraphics[width=0.5\linewidth]{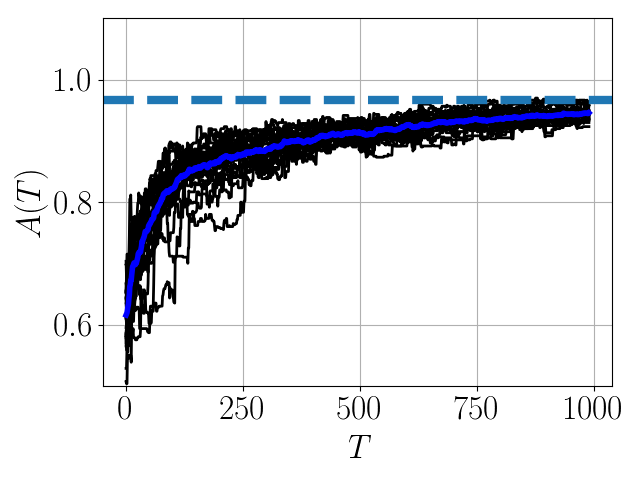}
		\includegraphics[width=0.5\linewidth]{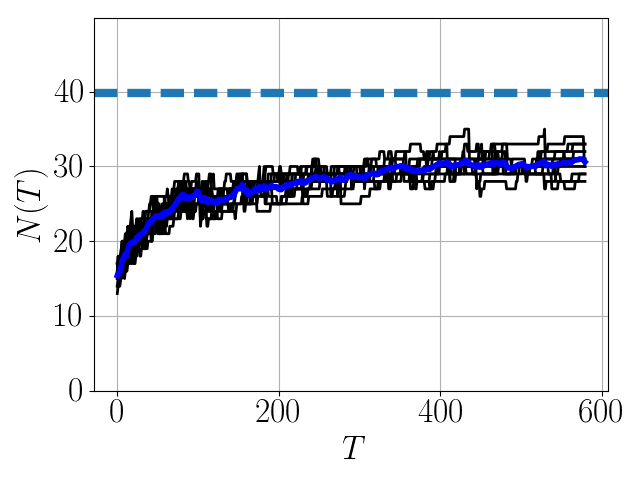}
	\end{tabular}

	\caption{Same as Figure \ref{linearFIG}, but for the non-linearly separable toy model, with true decision boundary $B_2$ defined in the SI text. The dimension is $p=30$ and the initial number of points is $N_{s}=30$.  }
	\label{nonlinearFIG}
\end{figure}

Figure \ref{linearFIG} and  \ref{nonlinearFIG} show that in both the linear and non-linear case for large $T$ the online algorithm converges to an accuracy $A(T)$ that is just below the accuracy of a batch SVM. Furthermore the number of active support vectors that the the online method finds after training is slightly below the number of support vectors that the batch SVM has.

\begin{figure}[h!]
	\begin{tabular}{ll}
		\includegraphics[width=0.5\linewidth]{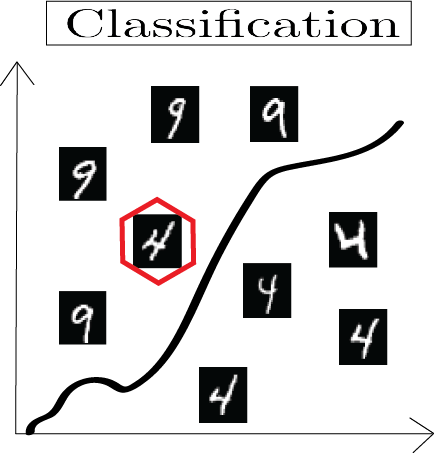}
		\includegraphics[width=0.5\linewidth]{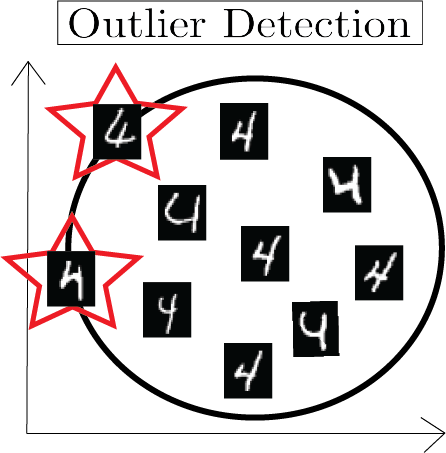}
	\end{tabular}
	\caption{Schematic showing instances of a supervised classification problem (left) and an outlier detection problem (right). In the supervised classification problem the goal is to partition the space into two distinct volumes, one for each label. The thick black line denotes the decision boundary. An incorrectly classified four (inside the red hexagon) is shown on the wrong side of the decision boundary. In the outlier detection problem, a sphere is created in the feature space to enclose the minimum volume while still containing all data points (in this case fours). Points on the boundary of this sphere are called outliers, they are starred in the schematic. The thick black line denotes the sphere boundary.   }
	\label{Computational_Problems}
\end{figure}

\section{Ecology to SVDD}\label{Ecology_to_SVDD}

We show how the method presented in the main text can be used to derive an approximate online SVDD learning algorithm first constructed by Jiang et al. in 2017 \cite{SAS_18}.

The Support Vector Data Description (SVDD) problem is concerned with unsupervised location of outliers in single-class classification problems (see \cite{Scholkopf_1999,Tax_2004} for a good overview). The SVDD problem consists finding a sphere of minimum radius in kernel space that contains all data points. Figure \ref{Computational_Problems} shows schematically a classification problem and an outlier detection problem. Points that lie on the surface of the sphere are called outliers and are analogous to active support vectors in the SVM problem. 

The problem is formulated as follows. Given a set of unlabeled data points $ \mathcal{D} = (x_{i}) _{i=1}^{N} $ : 
\begin{align}
\begin{split}
&\text{minimize } R^{2} \\
&\text{subject to } | \phi(x_{i}) - \mu  |^{2} \leq R^{2}  \text{ for all }  i
\end{split}
\end{align}
where $R$ is the sphere radius in feature space $\phi (x)$ and $\mu$ is the center of the sphere in the feature space.
This problem is simplified by the introduction of KKT multipliers $a_{i}$ for each inequality constraint. The KKT conditions for the minimization problem are:

\begin{align}\label{SVDDlag1}
\begin{split}
& 0 = \nabla_{R,\mu} L(R,\mu,a_{i}) \\
& 0 \leq a_{i} \\
& | \phi(x_{i}) - \mu  |^{2} -  R^{2} \leq 0 \\
& a_{i}[ | \phi(x_{i}) - \mu  |^{2} -  R^{2}]  = 0 \\
\end{split}
\end{align}

with 
\begin{equation*}
L(R,\mu,a_{i}) = R^{2} + \sum_{i=1}^{N} a_{i}( | \phi (x_{i}) - \mu |^{2}  - R^{2}  )\\
\end{equation*}

Minimizing $L(R,\mu,a_{i})$ with respect to $\mu$ and $R$ gives equations:
\begin{eqnarray}\label{SVDDmin}
\sum_{i=1}^{N} a_{i} = 1 \quad \text{and} \quad \mu = \sum_{i=1}^{N} a_{i} \phi (x_{i})
\end{eqnarray}
Substituting Equation \eqref{SVDDmin} into Equation \eqref{SVDDlag1} gives maximization problem for the optimal $a_{i}$:
\begin{align}\label{SVDDminimization}
\begin{split}
&\text{argmax}_{a_{i}} \text{ }L(a_{i}) = \sum_{i=1}^{N} a_{i} K(x_{i},x_{i}) -  \sum_{i,j=1}^{N} a_{i}a_{j} K(x_{i},x_{j})   \\
& \text{subject to }0 \leq a_{i} \text{ for all }i  \\
&\text{and } \sum_{i=1}^{N}  a_{i}  = 1
\end{split}
\end{align}
$L(a_{i})$ is called the dual SVDD Lagrangian and $K(x_{i},x_{j}) = \phi(x_{i})^{T} \phi(x_{j}) $. For simplicity we set $K(x_{i},x_{i}) = 1$ for the diagonal elements in the rest of the derivation, although the results we present are easily generalized for arbitrary kernel. The Python code included in the supplemental material works for any choice of $K(x,y)$.

After the SVDD is trained, the sphere radius in feature space $R$ can be determined via
\begin{equation}\label{SVDDradius}
\begin{split}
&R^{2} = \text{max}_{i} \text{ }  | \phi (x_{i}) - \mu |^{2}  \\
&  = \text{max}_{i} \text{ }( \phi (x_{i}) - \mu )^{T} ( \phi (x_{i}) - \mu ) \\
& =\text{max}_{i} \text{ } ( \phi (x_{i})^{T}\phi (x_{i}) - 2 \mu^{T} \phi(x_{i}) + \mu^{T} \mu ) \\
\end{split}
\end{equation}
The explict dependence on $\phi(x)$ in \eqref{SVDDradius} can be removed using $\mu = \sum_{i=1}^{N} a_{i} \phi (x_{i})$ :
\begin{equation*}
\begin{split}
& \phi (x_{i})^{T}\phi (x_{i}) - 2 \mu^{T} \phi(x_{i}) + \mu^{T} \mu   = \\ 
& 1 - 2 \sum_{j=1}^{N} K(x_{i},x_{j}) a_{j} + \sum_{j,k=1}^{N} K(x_{j},x_{k})a_{j}a_{k}
\end{split}
\end{equation*}
where we have used $\phi(x_{i})^{T} \phi(x_{j}) = K(x_{i},x_{j})$ and $K(x_{i},x_{i})=1$.
Thus, \eqref{SVDDradius} can be written in terms of kernel function and support vectors as
\begin{equation}\label{SVDDradius2}
\begin{split}
&R^{2} = \text{max}_{i} \big[  \\
&  1 - 2 \sum_{j=1}^{N} K(x_{i},x_{j}) a_{j} + \sum_{j,k=1}^{N} K(x_{j},x_{k})a_{j}a_{k}  \big]
\end{split}
\end{equation}

We apply our method to the SVDD Lagrangian \ref{SVDDminimization}. As $K(x_{i},x_{i})=1$ and $\sum_{i=1}^{N} a_{i} = 1$ the first term in the sum can be ignored. A Lagrange multiplier $\lambda$ is introduced to enforce the latter constraint. The quantity to be maximized is then
\begin{align}\label{SVDDlag}
\begin{split}
&\text{argmax}_{a_{i} , \lambda} L(a_{i},\lambda) = - \sum_{i,j=1}^{N} a_{i}a_{j}K(x_{i},x_{j}) +  \lambda ( \sum_{i=1}^{N}  a_{i}  - 1)   \\
& \text{subject to }0 \leq a_{i} \text{ for all }i \\
\end{split}
\end{align}

Using the quadratic programming-ecology duality, we can embed the solution to \eqref{SVDDminimization} as the steady state of the dynamical equations
\begin{align}\label{SVDDpde}
\begin{split}
&\frac{d a_{i}}{dt} = a_{i} (   \lambda - \sum_{j=1}^{N} K(x_{i},x_{j})  a_{j} ) \\
&\frac{d \lambda}{dt} =  1 - \sum_{i=1}^{N} a_{n}   
\end{split}
\end{align}

Now, suppose we are at the steady state of Equation \eqref{SVDDpde} and consider the addition of a new point $x_{0}$. 
The invasion condition is that the initial growth rate is positive
\begin{equation}\label{SVDDgrowth}
0 < \frac{1}{a_{0}} \frac{d a_{0}}{dt} =    \lambda - \sum_{j=1}^{N} K(x_{0},x_{j})  a_{j}  
\end{equation}
Let $x_{k}$ be any point which has non-zero support vector, $a_{k}>0$. Then, solving the steady state equation \eqref{SVDDpde} for auxiliary variable $\lambda$ gives
\begin{equation}\label{SVDDlambda}
\lambda =    \sum_{i=1}^{N} K(x_{i},x_{k})a_{i} 
\end{equation}
Inserting this into Equation \eqref{SVDDgrowth} gives the invasion condition
\begin{equation}\label{SVDDinvasion}
0 < \frac{1}{a_{0}} \frac{da_{0}}{dt} =  \sum_{i=1}^{N} a_{i}[K(x_{k},x_{i}) - K(x_{0},x_{i})]
\end{equation}
which is identical to Equation (2.9) derived in \cite{SAS_18} (Note that they use notation $z$ for our variable $x_{0}$). Equation
\eqref{SVDDinvasion} can be used to formulate an online learning algorithm in the same manner as the EcoSVM algorithm presented in the main text.
The paper by Jiang et al. derives Equation \eqref{SVDDinvasion} and calls the algorithm Fast Incremental Support Vector Data Description (FISVDD). They also numerically shows that this algorithm performs well on real-world datasets. The fact that this algorithm can be constructed simply and elegantly using a duality between quadratic programming and ecology suggests that there is a deeper connection between machine learning and ecological dynamics than previously realized and, more significantly, that ecologically inspired machine learning models are not just of theoretical interest but can be used for real world data analysis.

\begin{figure}[t!]
	\includegraphics[width=1.0\columnwidth]{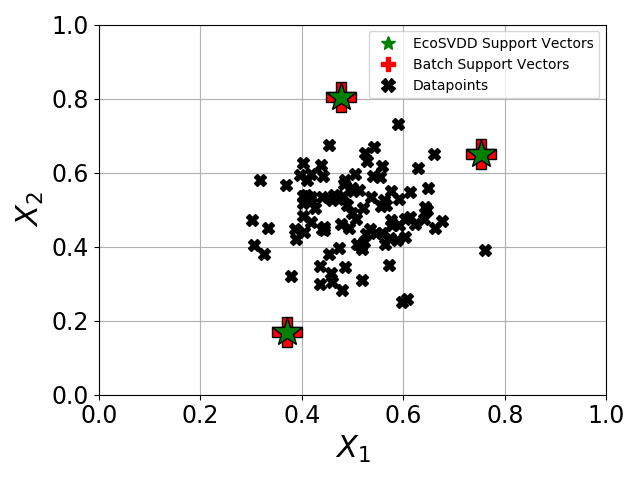}

	\caption{Comparison of batch SVDD and online SVDD algorithms for a data set with $N=100$ total points (shown in black) drawn from a Gaussian distribution. Batch SVDD active support vectors are shown with red ``+" symbols. EcoSVDD active support vectors are shown with green stars. The kernel function is Gaussian $K(x,y) = \exp( \frac{-1}{2}( x - y )^{T}(x - y  ) ) $. The EcoSVDD algorithm was started with $10$ points.  } 
	\label{SVDDfig1}
\end{figure}

We illustrate the ability of FISVDD/EcoSVDD numerically.  We draw data points from a $p$-dimensional multinomial Gaussian distribution with identity covariance matrix and mean uniformly sampled from the $p$-dimensional hypercube $ x \in [0,1]^{p}$. Figure \eqref{SVDDfig1} shows the two-dimensional case.

We define $R(T)$ to be the SVDD radius \eqref{SVDDradius2} of an FISVDD/EcoSVDD trained on $T$ points. The left panel of Figure \ref{SVDDfig1} shows $R(T)$ as a function of $T$. $R(T)$ converges to the batch SVDD radius (shown with a dashed blue line).

Similarly, we define a similarity metric between the FISVDD/EcoSVDD kernel sphere center trained on $T$ points $\mu(T)$ \eqref{SVDDmin} and the batch SVDD sphere center $ \tilde{\mu}$ as 
\begin{equation}
S(T)  = \frac{\mu(T)^{T}  \tilde{\mu} }{  \sqrt{ \mu(T)^{T}\mu(T)  } \sqrt{  \tilde{\mu}^{T} \tilde{\mu}  }  }
\end{equation}

\begin{figure}[ht]
	
	\begin{tabular}{ll}
		\includegraphics[width=0.5\linewidth]{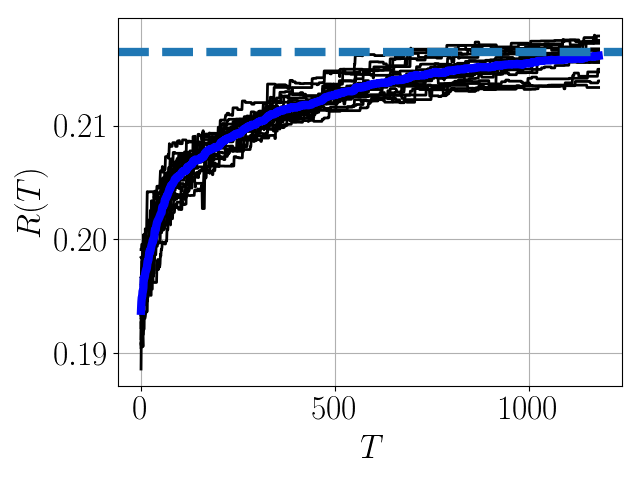}
		\includegraphics[width=0.5\linewidth]{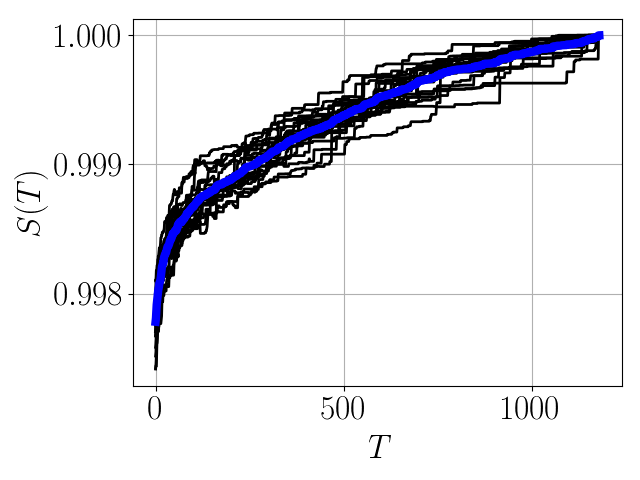}
	\end{tabular}

	\caption{Online SVDD radius and kernel sphere center similarity score as function of $T$. Left panel shows online SVDD radius $R(T)$ as a function of the number of points $T$ that the online SVM has seen. Black lines show individual realization, blue line shows mean radius. Dotted blue line shows batch SVDD radius. Right panel shows the normalized dot product between $\mu(T)$ and $\tilde{\mu}$.
	Black lines show individual realizations, blue line shows average over realizations. The dimension of the data space is $p = 15$ and the online training is initialized with $N_{s} = 30$ data points. }
	\label{SVDDfig2}
\end{figure}

$S(T) \leq 1$ with equality if and only if $\mu(T) = \tilde{\mu}$. The right panel of Figure \ref{SVDDfig2} shows $S(T)$ as a function of $T$. $S(T)$ converges to $1$ illustrating the fact that the FISVDD/EcoSVDD and batch SVDD produce the same kernel sphere center and radius.

\newpage
\bibliographystyle{apsrev4-1}
\bibliography{./eco2.bib}

\end{document}